\documentclass{article}
\usepackage{arxiv}

\usepackage[utf8]{inputenc} 
\usepackage[T1]{fontenc}    
\usepackage{hyperref}       
\usepackage{url}            
\usepackage{booktabs}       
\usepackage{amsmath}        
\usepackage{amsfonts}       
\usepackage{microtype}      
\usepackage{cleveref}       
\usepackage{graphicx}
\usepackage{natbib}
\usepackage{doi}

\title{Toward Reducing Unproductive Container Moves: Predicting Service Requirements and Dwell Times}

\author{ \href{https://orcid.org/0009-0005-6389-291X}{\includegraphics[scale=0.06]{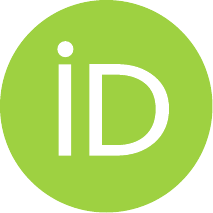}\hspace{1mm}Elena Villalobos} \\
	Centro de Ciencia de Datos e Inteligencia Artificial\\
    School of Government and Public Transformation\\
	Tecnológico de Monterrey\\
	Mexico City, Mexico\\
	\texttt{villalobos\_elena@tec.mx} \\
    \And
	\href{https://orcid.org/0000-0003-3499-0965}{\includegraphics[scale=0.06]{orcid.pdf}\hspace{1mm}Adolfo De Unánue T.} \\
	Centro de Ciencia de Datos e Inteligencia Artificial\\
    School of Government and Public Transformation\\
	Tecnológico de Monterrey\\
	Mexico City, Mexico\\
	\texttt{unanue@tec.mx} \\
    \And
    \href{https://orcid.org/0000-0001-8901-6022}{\includegraphics[scale=0.06]{orcid.pdf}\hspace{1mm}Fernanda Sobrino} \\
    Centro de Ciencia de Datos e Inteligencia Artificial\\
    School of Government and Public Transformation\\
    Tecnológico de Monterrey \\
	Mexico City, Mexico\\
    \texttt{fersobrinno@tec.mx} \\
    \And
    David Aké \\
    Centro de Ciencia de Datos e Inteligencia Artificial\\
    School of Government and Public Transformation\\
    Tecnológico de Monterrey \\
	Mexico City, Mexico\\
    \And
    Stephany Cisneros \\
    Centro de Ciencia de Datos e Inteligencia Artificial\\
    School of Government and Public Transformation\\
    Tecnológico de Monterrey \\
	Mexico City, Mexico\\
    \texttt{stephany.cisneros@tec.mx} \\
    \And
	  Jorge Lecona \\
    Container Terminal Operations \\
    Veracruz, Mexico \\ 
	\And
	Alejandra Matadamaz\\
    Container Terminal Operations \\
    Veracruz, Mexico \\ 
}

\renewcommand{\headeright}{}
\renewcommand{\undertitle}{}
\renewcommand{\shorttitle}{Toward Reducing Unproductive Container Moves: Predicting Service Requirements and Dwell Times}

\hypersetup{
pdftitle={Toward Reducing Unproductive Container Moves: Predicting Service Requirements and Dwell Times},
pdfsubject={},
pdfauthor={Elena Villalobos, Adolfo De Unánue T., Fernanda Sobrino, David Aké, Stephany Cisneros, Jorge Lecona, Alejandra Matadamaz},
pdfkeywords={Machine Learning,Port Terminal, Data Product},
}

\begin{document}

\maketitle

\begin{abstract}
    This article presents the results of a data science study conducted at a
    container terminal, aimed at reducing unproductive container moves through
    the prediction of service requirements and container dwell times. We
    develop and evaluate machine learning models that leverage historical
    operational data to anticipate which containers will require pre-clearance handling services prior to cargo release and to estimate how long 
    they are expected to remain in
    the terminal. As part of the data preparation process, we implement a
    classification system for cargo descriptions and perform deduplication of
    consignee records to improve data consistency and feature quality. These
    predictive capabilities provide valuable inputs for strategic planning and
    resource allocation in yard operations. Across multiple temporal validation
    periods, the proposed models consistently outperform existing rule-based
    heuristics and random baselines in precision and recall. These results
    demonstrate the practical value of predictive analytics for improving
    operational efficiency and supporting data-driven decision-making in
    container terminal logistics.
\end{abstract}


\section{Introduction}\label{sec:intro}

Container terminals play a central role in global supply chains, acting as
critical nodes where maritime and inland transportation systems converge. Their
efficiency directly affects vessel turnaround times, hinterland connectivity,
and the overall performance of international trade networks \citep{UNCTAD2025RMT}.
As global cargo volumes continue to grow, optimizing terminal operations has become increasingly important for maintaining competitiveness and ensuring service reliability.

Within container terminals, yard management is one of the most complex and
consequential operational processes. Decisions regarding where and how
containers are stacked influence equipment utilization, operational costs, and
the number of unproductive moves—such as reshuffles (rehandling moves)—required to
access containers buried beneath others \citep{gharehgozliHeuristicEstimationContainer2017}.
Reducing these non-productive moves is
essential, as they account for a significant share of total handling activity
and represent substantial resource consumption, including fuel, labor time, and
equipment wear.

Traditional approaches to yard planning rely on deterministic optimization
models or static heuristics that assume known departure sequences. However,
container behavior in practice is inherently stochastic: certain containers
require pre-clearance processes prior to departure, consignee pickup
behavior follows probabilistic rather than fixed schedules, and dwell times are
influenced by complex interactions among cargo type, shipping line, origin
port, and seasonal factors. These characteristics make the problem well-suited
to machine learning \citep{bengioMachineLearningCombinatorial2018},
which can exploit large volumes of historical operational
data to learn predictive patterns that would be difficult to encode manually.
Rather than optimizing a stacking plan given assumed departures, we reframe the
problem as one of \emph{prediction}.

This study develops a data-driven system designed to support yard planning
by estimating two key pieces of information before
container arrival: whether a container will require a pre-clearance handling
service, and how long it is expected to remain in the terminal.
Pre-clearance services refer to administrative procedures conducted by
customs brokers prior to cargo release, which require the terminal to
reposition the container for handling. By anticipating these service
requirements and dwell times, yard planners can make more strategic
stacking decisions, placing containers requiring service closer to
designated areas, short-stay containers in more accessible locations,
and long-stay containers in lower-priority zones, thereby reducing
unnecessary reshuffles.

\section{Literature review}\label{sec:lit}

Research on container terminal operations has evolved from rule-based heuristics
to data-driven approaches that leverage machine learning to improve yard
efficiency. This section reviews the literature, organized around three themes:
container dwell-time prediction, stacking and relocation optimization, and the
integration of predictive analytics in terminal operations.

\subsection{Container Dwell Time Prediction}

The prediction of container dwell time has received increasing attention as
terminals seek to optimize yard space
utilization. \citet{kourouniotiDevelopmentModelsPredicting2016} proposed one of
the first systematic approaches using Artificial Neural Networks (ANNs) to
predict import container dwell times, identifying discharge timing, port of
origin, container dimensions, and cargo type as key determinants.

More recent studies have explored ensemble machine learning methods.
\citet{yoonComparativeStudyMachine2023} compared six algorithms—including Random Forest,
XGBoost, and LightGBM—for vessel dwell-time prediction at Busan Port, finding
that all ML models outperformed the terminal's operational reference.
\citet{sainiASSESSINGFACTORSIMPACTING2024} analyzed 2.8 million container records across
fourteen ports, identifying factors such as free storage periods, transshipment
status, and proximity to industrial hubs as significant predictors of dwell-time
variation.

The need for model interpretability has led to research incorporating
Explainable AI (XAI). \citet{leeIdentifyingKeyFactors2024} combined Process Mining with
XAI techniques to identify key factors prolonging container dwell times,
emphasizing the importance of transparency for operational adoption.

\subsection{Container Stacking and Reshuffles}

The container stacking problem is fundamental to yard efficiency.
\citet{kaphwankimEvaluationNumberRehandles1997} established the mathematical
foundation for quantifying rehandle costs based on stacking configurations,
demonstrating that stack height and departure sequence uncertainty are primary
drivers of unproductive moves.

Various optimization approaches have been proposed to minimize reshuffling.
\citet{chafikStackingPolicySolving2016} developed a Mixed Integer Programming (MIP) model
comparing First-Come-First-Served and Best Fit Decrease heuristics.
\citet{casertaApplyingCorridorMethod2011} applied metaheuristic methods to the block
relocation problem, demonstrating scalability to realistic terminal sizes.
\citet{borgmanOnlineRulesContainer2010} proposed online stacking rules for real-time
decisions, highlighting that accurate departure time information significantly
improves stacking outcomes.

The integration of prediction with optimization represents an emerging research
direction. \citet{gaeteg.DwellTimebasedContainer2017} proposed a decision support system that
first predicts dwell times using Random Forest regression and then applies
heuristics to minimize reshuffles-an architecture closely related to the present
study.

\subsection{Machine Learning in Maritime Operations}

The broader adoption of machine learning in port operations has been documented
in recent systematic reviews. \citet{jahangardLeveragingMachineLearning2025} analyzed 124
papers on predictive and prescriptive analytics in seaports, identifying a gap
in research that combines predictive outputs with optimization models.
\citet{heiligDigitalizationDataDrivenDecision2019} established a conceptual framework for
data-driven terminal planning, emphasizing that analytics reduces uncertainties
and identifies causes of operational inefficiencies.

Related prediction problems in port logistics demonstrate the versatility of ML
approaches. \citet{xiePredictingOutterminalsImported2025} applied machine learning to predict
container exit terminals, incorporating unstructured cargo descriptions.
\citet{saberHighaccuracyPredictionVessels2025} developed hybrid ML models for vessel arrival
prediction, demonstrating the value of combining multiple algorithmic
approaches.

\subsection{Research Gaps and Contributions}

Despite substantial progress, several gaps remain in the literature. First,
while dwell time prediction has received considerable attention, the prediction
of \emph{service requirements} from a terminal operator's perspective remains
largely unaddressed. Substantial research exists on customs risk assessment for
enforcement purposes—including ML-based targeting systems that help authorities
identify containers for inspection based on contraband or compliance risks
\citep{hillberryRiskManagementBorder2022,vijayakumarTechnologycentricDataDrivenCustoms2025,PerspectiveRiskManagement2020}.
However, these models serve government agencies rather than terminal operators.
Predicting which containers will require pre-clearance handling for
yard planning purposes—where the goal is operational efficiency rather than
enforcement targeting—has not been explored. Second, most studies address either
prediction or optimization in isolation; integrated approaches that feed
predictions directly into operational decisions are limited
\citep{jahangardLeveragingMachineLearning2025}. Third, rigorous temporal
validation using frameworks such as temporal cross-validation \cite[see][]{robertsCrossvalidationStrategiesData2017} is rarely
implemented, raising concerns about data leakage \cite[see][]{rayidghaniTop10Ways2020} and deployment validity.
Fourth, comparisons typically use random or dummy baselines rather than actual
operational heuristics employed by terminals.

This study addresses these gaps by: (1) developing models for both service
requirement and dwell-time prediction; (2) employing rigorous temporal cross-validation\footnote{Employing the Triage framework
\citep{centerfordatascienceandpublicpolicyTriageGeneralPurpose2025,rayidghaniTriageMLData2024}}; (3) comparing
against operational baselines currently used in practice; and (4) providing
empirical results from a Latin American container terminal, expanding the
geographic scope of this research domain.

\section{Problem description}\label{sec:prob}

One of the critical operational processes in container terminals is the
placement of import containers in the yard area. Each day, hundreds of
containers are unloaded from incoming vessels and distributed across different
yard blocks after discharge. The movement of containers within the terminal---from
the quay to storage areas and eventually to the gate---is referred to as a
\textbf{container move}. To optimize the use of available space, terminals stack
containers vertically, a practice known as \textit{container stacking}.

The stacking of containers inevitably leads to situations in which containers
scheduled for departure or service are located beneath others. Retrieving
these buried containers requires temporarily removing those above them, a process
referred to as \textit{reshuffling} or \textit{rehandling}. Terminal operators
classify these operations as \textbf{unproductive moves} or \textbf{waste},
since they do not directly contribute to the logistical flow of cargo.

Unproductive moves represent one of the major sources of inefficiency in yard
operations. In the terminal where this study was conducted, up to 75\% of all
container-handling moves were classified as unproductive. Of these, approximately
51\% were associated with containers requiring a pre-clearance service (hereafter
referred to as ``service''), i.e., an administrative process carried out by
customs brokers prior to cargo release, which requires the terminal to perform
additional handling operations to position the container for review\footnote{This process
should not be confused with formal customs inspections conducted by authorities,
which fall outside the scope of this study.}. These figures indicate that containers
requiring pre-clearance handling are a particularly significant driver of
operational inefficiency in this setting.

A promising strategy to mitigate these inefficiencies is to anticipate,
\emph{before a vessel is discharged}, which containers are likely to require
a service and to estimate their expected dwell time. Early
identification of containers likely to require pre-clearance services enables yard
planners to position them closer to designated service areas, reducing the distance
and number of handling operations required for subsequent retrieval. Similarly,
predicting dwell time supports differentiated stacking: containers expected to
leave the terminal sooner can be placed in more accessible positions, while
long-stay containers can be stored in less critical sections of the yard without
obstructing outbound flows.

By integrating predictive information on both service and dwell
times, yard planners can make more strategic and proactive placement decisions.
This approach has the potential to substantially reduce rehandling operations,
improve equipment utilization, and optimize overall yard space management. To
operationalize this predictive framework, the next section presents the data
sources, operational context, and system architecture that support the
development and evaluation of the proposed models.

\section{Data, operational context and architecture}\label{sec:dop}

\subsection{Data}\label{sub:datos}

The port terminal provided two databases: (1) container operations and (2)
container movements within the terminal yard. Additionally, a publicly available
database known as the HS Catalog (\textit{Harmonized System}) was used. A brief
description of each dataset is presented below.

\begin{description}
\item[Operations:] The terminal generated a view\footnote{Views are virtual
      tables formed by a \emph{query} that are computed each time they are
      called.} from its CTS production database.\footnote{CTS stands for
      \textit{Container Terminal System}, the production database used by the
      terminal.} To create this view, the terminal combined information from
    various tables referring to container attributes and date-related
    information. The original version of the view contains approximately two
    million containers (a similar number to the work of
    \citet{sainiASSESSINGFACTORSIMPACTING2024}), covering the most recent
    several years of operations\footnote{Exact calendar years are omitted at the request of the terminal port to preserve confidentiality. Temporal patterns are presented using relative time references and anonymized year labels (e.g., ``20XX'').}. Some examples of the container attributes
    included in this view are: loading port, container dimensions and type,
    cargo description, consignee, vessel arrival date, container entry and exit
    dates, country of origin, among others.
\item[Movements:] The terminal also shared another view from its database
    containing the movements performed by cranes to relocate containers within
    the terminal yard. This dataset contains approximately 13 million rows. The
    data spans the same multi-year period and was used for exploratory analysis
    of container movements within the terminal.
\item[HS Catalog:] The HS Catalog (\textit{Harmonized System}) is publicly
    available and serves as a numerical standardization method for the
    classification of traded goods. It is used to identify products for tariff
    and tax assessment and for the compilation of trade statistics. The catalog
    is updated every five years and serves as the foundation for international
    import and export classification systems. Specific six-digit codes are
    assigned to different categories and
    products\footnote{\url{https://www.trade.gov/harmonized-system-hs-codes}.}.
\end{description}

Access to operational data was subject to strict security and governance
controls. All preprocessing and initial data transformations were conducted
within the terminal’s own computing infrastructure. Dedicated machines were
provisioned with restricted permissions, allowing access only to the specific
container attributes required for predictive modeling. Sensitive fields were
excluded whenever possible and, when operationally necessary, were encrypted or
anonymized by the terminal prior to researcher access. No direct access was
granted to core production systems, and all analytical workflows operated on
controlled extracts and derived views. These measures ensured compliance with
the terminal’s data protection policies while preserving the analytical value of
the datasets.

\subsection{Data Product Pipeline}\label{dataproduct}

\begin{figure}
    \centering
    \includegraphics[width=1\textwidth]{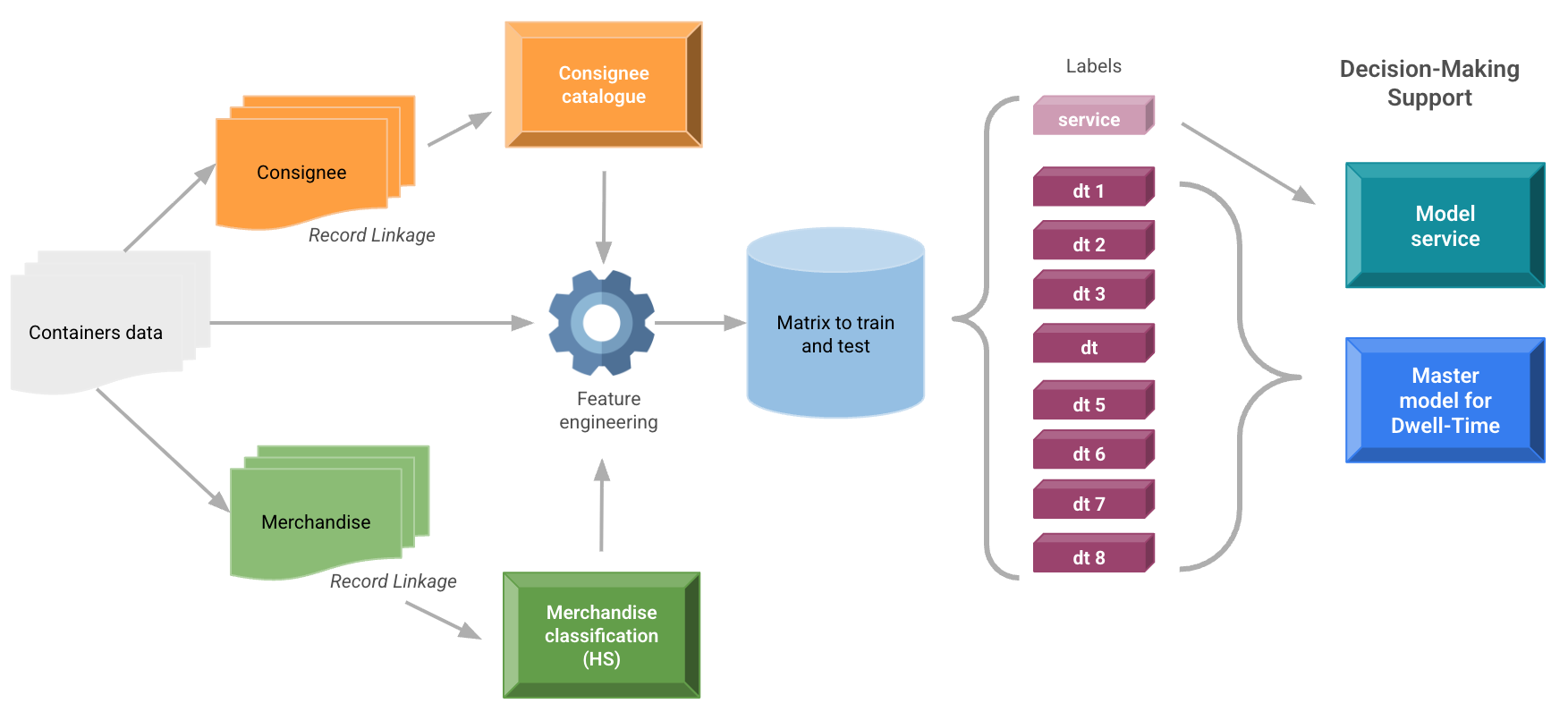}
    \caption{Data product pipeline.}
    \label{fig:pipeline}
\end{figure}

To predict container service requirements and dwell times (DT), we developed a
comprehensive data processing and enrichment pipeline. This workflow forms the
technical backbone of the data product that supports predictive modeling for
enhanced operational yard planning (see Figure \ref{fig:pipeline}).

The pipeline starts with the integration of heterogeneous \textbf{operational data}
sources generated across different systems involved in container terminal
operations. To ensure traceability and reproducibility, the data are organized
into a staged schema architecture. First, all incoming datasets are ingested
as-is into a schema referred to as \textsc{raw}, which preserves the original
structure, formats, and potential inconsistencies of the source systems.

In a second step, the data are transformed into a \textsc{clean} schema, where
variables are standardized, cast to their appropriate data types, and
semantically interpreted according to their operational context. This stage
resolves inconsistencies, harmonizes naming conventions, and establishes a
coherent representation of the underlying processes.

For the modeling stage, a domain-driven data model is constructed in a schema
named \textsc{ontology}. This schema is organized around two main structures.
The first corresponds to entities, which in this study are containers and their
atemporal attributes—i.e., characteristics that remain invariant over time,
such as weight, dimensions, or cargo type. The second structure captures events,
representing all operational occurrences that affect entities over time. These
events are explicitly associated with timestamps, enabling a temporal
representation of container trajectories and operational states.

Subsequently, two record linkage processes are implemented to enrich the data model and ensure consistency across operational entities. The first focuses on the classification of merchandise descriptions into standardized cargo categories, while the second addresses the deduplication and consolidation of consignee identities. Both processes, briefly described below, are essential for generating stronger predictors and enabling reliable modeling.

\textbf{Merchandise:} The database includes a variable denoted as merchandise description, which is
a free-text field provided by users to describe the contents of each container.
These descriptions may consist of natural-language text or, in some cases,
references to codes derived from the Harmonized System (HS) catalog. Due to this
heterogeneity, the raw descriptions cannot be directly used as structured
predictors.

To obtain a standardized representation of cargo type, merchandise
classification is grounded on the Harmonized System (HS) nomenclature, which
provides an internationally recognized taxonomy comprising 21 high-level
sections and 97 detailed chapters. However, only approximately 25\% of the
containers include an explicit HS reference in their description. Consequently,
an automated classification approach based on Natural Language Processing (NLP)
is required for the remaining cases.

Specifically, a TF–IDF representation \citep{saltonTermweightingApproachesAutomatic1988} is employed to quantify
the relevance of terms in each merchandise description relative to the HS
catalog. TF–IDF vectors are constructed for HS chapters using their textual
definitions and compared against container-level descriptions to assign the
most relevant chapter. This mapping yields a standardized cargo category that is
subsequently used as a predictive feature across all models.

Alternative embedding-based approaches, including Word2Vec
\citep{mikolovDistributedRepresentationsWords2013}, were evaluated but ultimately discarded due to higher
computational cost and inferior classification performance in this setting.
The selected TF–IDF-based approach was validated through manual inspection of a
random sample of container descriptions and qualitative comparison between
HS chapter definitions and the resulting classified descriptions. As a result of this process, standardized merchandise classification coverage
increased to 88\% of all containers.


\textbf{Consignee:} This information contains variations in spelling, abbreviations,
punctuation, or formatting for
the same underlying entity. These inconsistencies lead to artificial inflation
in the number of unique consignees and introduce noise when the variable is used
directly for analysis or modeling. To address this issue, a record linkage
process is implemented to deduplicate and consolidate consignee identities.

The linkage procedure begins by generating candidate pairs of consignees and
computing their string similarity using character-based trigrams. To reduce the
computational burden associated with exhaustive pairwise
comparisons, a blocking strategy is applied. Comparisons are restricted to
consignees whose names share the same initial character, which substantially
reduces the number of candidate pairs while preserving relevant matches.

Candidate pairs with similarity scores above a threshold of 0.8\footnote{The threshold was selected based on manual inspection of candidate pairs across different similarity levels.} were retained 
and represented as edges in an undirected graph, where nodes correspond to
consignee identifiers. A depth-first search (DFS) algorithm is then applied to
identify connected components within the graph, effectively grouping
the different textual representations of the same consignee across the dataset.

This graph-based consolidation process results in a reduced and more consistent
consignee catalog, where each connected component represents a unique underlying
entity. By resolving duplicate identities and harmonizing consignee information,
this procedure improves data consistency and reduces noise in downstream
analyses. The resulting consolidated consignee identifier is subsequently used
as a predictive feature in the models described in the following section.

Both standardized variables, the consolidated merchandise category and the
deduplicated consignee identifier, are incorporated into the feature engineering
stage. This stage is designed to derive explanatory variables that capture both
operational dynamics and service-related patterns observed throughout the
container lifecycle. The outcome is a structured \textbf{feature matrix} used for
training and validating the machine learning models. This matrix is associated with \textbf{two labeling schemes}: one identifying whether a container will require service, and another specifying the expected dwell-time interval within the terminal.

Dwell time labels are further refined through a \textbf{decision rule} that assigns a
unique final label to each container. Consequently, the data product outputs two
ranked lists based on model scores. The first ranking orders containers
according to their likelihood of requiring service, while the second ranking is
derived from a decision rule applied to the dwell-time model scores, as
described in the following sections.

\section{Methodology}

\subsection{Analytical Formulation}\label{sub:labelscohorts}

This section describes the main elements used for model evaluation: the cohort and the labels. The cohort refers to the set of entities on which predictions are made, while the labels define the outcomes to be predicted.

\begin{description}
    \item[Cohort:] Containers that are within 24 hours of arriving at the terminal.
    \item[Labels:] Nine distinct labels were defined, each determined 24 hours prior to the vessel’s arrival at the terminal:
\end{description}

\begin{enumerate}
    \item Which are the $n$ containers that will leave the terminal in less than two days?
    \item Which are the $n$ containers that will leave the terminal on day $m$?, where $m \in \{2,3,4,5,6,7\}$
    \item Which are the $n$ containers that will leave the terminal after more than seven days?
    \item Which are the $n$ containers that will receive service within the next seven days?
\end{enumerate}

In earlier stages of the project, a more aggregated classification scheme was used: \textit{less than two days}, \textit{two to four days}, \textit{four to seven days}, and \textit{more than seven days}. However, a finer-grained, day-level labeling scheme was ultimately adopted to enhance predictive performance and generate more operationally actionable insights.

The choice to train separate binary classifiers for each dwell-time label—rather than a single multi-class model or regression approach—was motivated by three considerations. First, binary decomposition enables label-specific threshold optimization via ROC analysis, allowing operational trade-offs between precision and recall to be calibrated independently for each dwell time category based on their distinct operational costs. Second, the predictability of dwell time labels varies substantially: extreme categories (short and long stays) exhibit stronger predictive signals than intermediate durations, and binary classifiers allow each model to exploit the feature space differently without forcing shared representations. Third, this architecture provides operational flexibility, as terminals may prioritize accurate identification of specific categories (e.g., short-stay containers for accessible positioning) while tolerating lower precision on others.

\subsubsection{Baseline Rate}

Understanding the \textbf{baseline rate} is fundamental for assessing the difficulty of the predictive problem. Based on the most recent years of historical data, if a random set of containers is selected on any given day, only about 33\% of them require at least one service.

Regarding dwell time, Table~\ref{tab:baserate} presents the baseline distribution of containers according to the defined labels. For instance, \textit{Dwell Time Label 1} corresponds to containers that exited the terminal in less than two days, whereas \textit{Dwell Time Label 8} groups those that remained for more than seven days. Most labels account for between 8\% and 12\% of cases, except for the last category, which represents 28\% of containers.

\begin{table}[htbp]
\centering
\caption{Baserate dwell-time distribution over the study period.}
\begin{tabular}{cll}
\toprule
Dwell Time Label & Cumulative Percentage & Label Percentage \\
\midrule
1 & 8\%  & 8\% \\
2 & 18.36\% & 10\% \\
3 & 30.60\% & 12\% \\
4 & 42\% & 12\% \\
5 & 54.07\% & 12\% \\
6 & 64.27\% & 10\% \\
7 & 72\% & 8\% \\
8 & $>$72\% & 28\% \\
\bottomrule
\end{tabular}
\label{tab:baserate}
\end{table}

\subsubsection{Baselines}

The terminal currently employs simple operational heuristics to anticipate which
containers will require pre-clearance services. These rules constitute the \textbf{current operational baseline}, and they
represent the level of performance that our machine learning models aim to
surpass.

\textbf{Operational Baseline 1.} This rule relies on the historical behavior of
consignees. Consignees that have frequently required pre-clearance services in the
past are assumed to have a higher probability of requiring service for future
containers.

\textbf{Operational Baseline 2.} Under this rule, if a consignee has required
pre-clearance services for more than 85\% of its containers over the previous six
months, it is assumed that upcoming containers associated with that consignee
will also require service.

In contrast, there is currently no automated mechanism or established
operational rule for predicting container dwell time. As a result, dwell time
models are evaluated against a \textbf{random assignment baseline}, which serves
as the reference performance they must improve upon.

\subsection{Features}

Feature selection was guided by two criteria: (i) variables had to be available prior to the prediction event, and (ii) they needed to capture meaningful operational information relevant to container behavior. A joint effort was undertaken to construct a detailed temporal mapping of all variables in the database, allowing us to identify the precise operational moment at which each variable was recorded. This temporal alignment was critical for the design of predictive experiments and for the correct implementation of Temporal Cross-Validation (see Subsection~\ref{sub:temporal}).

\paragraph{Base variables.}
Predictor variables were derived from information available prior to container arrival at the terminal. Table~\ref{tab:base_features} summarizes the core variables that define the foundation of the feature space. Variables marked with an asterisk (*, **) underwent additional \emph{record linkage} processes (see Section~\ref{dataproduct}) to standardize consignee entities and to assign structured classifications (chapter and section) to merchandise descriptions.

\begin{table}[htbp]
\centering
\caption{Base variables available prior to container arrival.}
\begin{tabular}{ll}
\toprule
\multicolumn{2}{c}{\textbf{Variable}} \\
\midrule
Net weight & Container dimension \\
Gross weight & Container type \\
Hazardous cargo indicator & Cargo type \\
Liner client (yes/no) & Shipping line \\
Consignee* & Shipping line route \\
Chapter and section** &  \\
\bottomrule
\end{tabular}
\label{tab:base_features}
\end{table}

\paragraph{Feature generation.}
From these base variables, an expanded feature set was constructed through systematic transformations, aggregations, and temporal windowing strategies. These operations were designed to capture behavioral regularities, temporal dynamics, and cross-entity relationships among operational actors (e.g., consignees, shipping lines, and cargo chapters).

Feature engineering included rolling-window statistics, historical frequency measures, trend indicators, and ratio-based variables computed at different aggregation levels. The following examples illustrate representative feature patterns used in the construction process (non-exhaustive):

\begin{itemize}
    \item Rolling service-frequency indicators, such as the number and proportion of containers requiring services for a given consignee over the past $n$ weeks.
    \item Rolling arrival counts and rates by cargo type or chapter within fixed temporal windows (e.g., three-week windows).
    \item Temporal trend features derived from recent dwell time distributions associated with specific shipping lines or routes.
\end{itemize}

In practice, multiple variants of these feature patterns were generated across different entities, time windows, aggregation levels, and normalization schemes, resulting in a rich and structured predictor space.

\paragraph{Feature taxonomy.}
To maintain a structured and reproducible framework, all predictors were systematically categorized according to their nature and derivation process. Table~\ref{tab:feature_categories} summarizes the taxonomy of generated features, which served as the foundation for model design and interpretability.

\begin{table}[htbp]
\centering
\caption{Feature categories and descriptions.}
\begin{tabular}{lp{10cm}}
\toprule
\textbf{Category} & \textbf{Description} \\
\midrule
Simple variables & Directly available from base container information, 24 hours prior to arrival. \\
Simple counts & Frequency of occurrence of specific variables within a defined time window. \\
Aggregated metrics & Ratios, proportions, and relationships between counts across temporal spans. \\
Differences & Temporal variations, percentage changes, and dispersion measures between periods. \\
Service & Frequency of services by consignee, chapter, or shipping line over time. \\
Dwell time & Descriptive statistics derived from historical container departure records. \\
Movements & Aggregated information on container movements by consignee and chapter. \\
\bottomrule
\end{tabular}
\label{tab:feature_categories}
\end{table}

\paragraph{Data integrity.}
To ensure predictive validity, strict controls were implemented to prevent
\textit{data leakage}. All features were computed exclusively from information
available before the prediction point, guaranteeing that models had no access to
future data. This approach ensures that model performance during training
faithfully reflects behavior under operational deployment.

\subsection{Model Governance}

A total of more than 5,000 models were trained under different configurations to
predict both service and dwell-time labels. Model training and data product
    generation were carried out using \texttt{Triage}, a predictive modeling
framework developed by the \emph{Data Science for Social Good} initiative
\citep{centerfordatascienceandpublicpolicyTriageGeneralPurpose2025}. Although originally designed for public policy
applications, \texttt{Triage} is particularly suited for data science projects
with a strong temporal structure, such as the present port terminal case study.

Model governance refers to the systematic oversight of decisions that directly
influence the predictive modeling process. It ensures reproducibility,
transparency, and methodological consistency throughout the experimental
pipeline. Within this project, governance was exercised across three key
dimensions: (i) the definition of predictive labels already described in Section
\ref{sub:labelscohorts}, (ii) the selection and parameterization of algorithms,
and (iii) the design of temporal configurations for training and validation.

\subsubsection{Algorithmic configuration}

A diverse ensemble of algorithms was explored to assess both linear and
non-linear modeling capacities, ranging from simple baselines to ensemble-based
classifiers. Table~\ref{app:dwell} in the Appendix summarizes the principal algorithms and
hyperparameters evaluated.

Model performance was evaluated against the operational baselines described in
Section~\ref{sub:labelscohorts}: consignee-based heuristics for service
prediction and random assignment for dwell time.

\subsubsection{Temporal Structure in Model Training and Validation}\label{sub:temporal}
Since the terminal data are inherently time-dependent, we implemented a
\textbf{Temporal Cross-Validation (TCV)} strategy
\citep{robertsCrossvalidationStrategiesData2017} for model training and
evaluation. This technique enables multiple experiments to be conducted under a
chronological structure consistent with real-world operations, adapting to the
specific nature of each label. In this approach, models are trained on
historical data and evaluated on future periods, thereby preserving the temporal
sequence and preventing information leakage (\textit{data leakage}).

In the present project, a temporal structure is defined using data spanning
three years of recent operations, which establishes the temporal bounds of the
analysis. Within this period, models are trained using rolling historical
windows of six months and subsequently evaluated over a one-month validation
period, which serves as a proxy for real-world deployment intervals. This design
allows the modeling framework to adapt to temporal trends and evolving
operational conditions.

Prediction points are generated on a daily basis, mirroring the operational
process in which containers arrive at the terminal each day. Outcome labels are
defined using rolling time windows whose length depends on the target variable.
Specifically, label occurrence windows range from 2 to 7 days, reflecting the
difference between short-horizon service requirements and longer dwell-time
dynamics.

Finally, all temporal configurations are updated through monthly model
retraining. This retraining frequency balances the need to incorporate recent
operational patterns while retaining sufficient historical information to
ensure stable and reliable learning.

\subsubsection{Decision Rule for Dwell Time Label Assignment}\label{sec:decision}

Since the dwell-time prediction problem is decomposed into eight binary
classification tasks, a decision rule is required to assign a single final
label to each container. This rule is informed by the analysis of the
\textit{Receiver Operating Characteristic} (ROC) curves generated for each
dwell-time model.

The ROC curve graphically represents the relationship between the \textit{True
Positive Rate} (TPR) and the \textit{False Positive Rate} (FPR) across
different classification thresholds. It serves as a standard tool to evaluate
the discrimination capacity of binary classifiers.

For each dwell-time label, the ROC curve was computed, and the optimal point was
identified—defined as the threshold that maximizes the difference between the
true positive rate and the false positive rate (also known as the Youden index).
This point was then used as the cutoff threshold to identify containers most
likely to belong to each label.

Since a single container could surpass the optimal threshold for multiple
labels, a rule was required to assign a unique dwell-time label to each
container. The adopted decision rule is based on the \emph{ranking of model
scores} assigned to each container. Specifically, among the labels for which a
container is classified as positive (i.e., surpassing the threshold), the label
with the highest relative score ranking is selected.

The procedure is as follows:

\begin{enumerate}
    \item For each label, containers with scores exceeding the optimal ROC-derived threshold are selected.
    \item Within each label, containers are sorted in descending order of their scores and assigned a ranking position.
    \item For containers appearing in multiple labels, the label corresponding to the best (highest) ranking is selected.
\end{enumerate}

This approach systematically resolves label overlaps by selecting the
classification where the model exhibits the greatest relative confidence.

\section{Results}\label{sec:results}

To evaluate model performance, two primary metrics were employed:
\textbf{precision} and \textbf{recall}, both calculated over the temporal
validation periods defined in the configuration. Precision measures the
proportion of true positives among the containers predicted as positive by the
model---of all containers the model identified as likely to
require service, how many actually did so? Recall represents the proportion of
actual positive cases correctly identified by the model---how
many of the containers that indeed required service were detected by the model?

These metrics were adapted to the operational constraints associated with each
predictive label. The following subsections present the results for
\textit{service prediction}, followed by those for \textit{dwell time}, each
with their specific analytical considerations.

Table~\ref{tab:key_results} summarizes the key findings across both prediction
tasks, comparing the best-performing models against their respective baselines.

\begin{table}[htbp]
\centering
\caption{Summary of key results across prediction tasks.}
\begin{tabular}{llccc}
\toprule
\textbf{Task} & \textbf{Best Model} & \textbf{Precision} & \textbf{Recall} & \textbf{vs. Baseline} \\
\midrule
Service ($k=300$) & Random Forest & 75\% & 100\% & +50 pp \\
\midrule
Dwell time $<$2 days & Random Forest & 30--40\% & 80\% & +50 pp \\
Dwell time 3--6 days & Random Forest & $<$25\% & $<$40\% & +15 pp \\
Dwell time $>$7 days & Random Forest & 80\% & 90\% & +40 pp \\
\bottomrule
\end{tabular}
\begin{flushleft}
\small\textit{Note:} Service baseline refers to operational heuristics; dwell-time baseline is random assignment. Metrics represent peak performance across validation periods. pp = percentage points.
\end{flushleft}
\label{tab:key_results}
\end{table}

\subsection{Service}

For service prediction, the model evaluation was grounded in the terminal’s
daily operational capacity to assign containers to the service
area. According to the terminal operations team, this capacity is approximately
300 containers per day. Accordingly, we set $k = 300$ to compute the metrics
\texttt{precision@k} and \texttt{recall@k}, defined as follows:

\begin{itemize}
    \item \texttt{precision@k:} proportion of correctly predicted containers among the top $k$ containers identified by the model as requiring service.
    \item \texttt{recall@k:} proportion of all containers that actually required service and were correctly identified within the top $k$ prioritized cases.
\end{itemize}

The value of $k$ can be adjusted to simulate different operational scenarios, as
illustrated later in this section.

\begin{figure}
\centering
\includegraphics[width=1\textwidth]{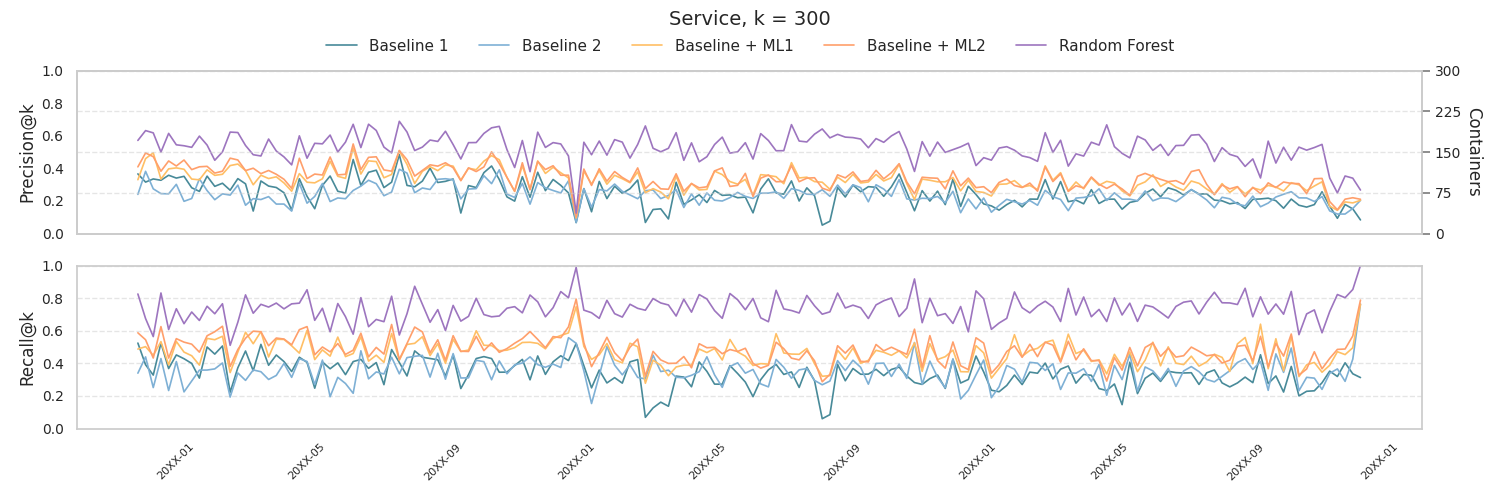}
\caption{Weekly average precision and recall for the service label across the main evaluated models. The upper panel displays \texttt{precision@k}, while the lower one shows \texttt{recall@k}.}
\label{fig:serv}
\end{figure}

Figure~\ref{fig:serv} summarizes the performance of the main models selected
from a broader set of evaluated approaches. While multiple models
were evaluated, the figure highlights representative results that reflect the
progressive analytical logic adopted in this study: starting from existing
operational heuristics, extending them with simple machine learning methods,
and ultimately deploying a fully data-driven model with enriched feature
engineering.

The evaluation begins with the two operational heuristics currently used by the
terminal to anticipate which containers will require service, represented in
blue as \texttt{Baseline 1} and \texttt{Baseline 2}. As mentioned before, these rules rely on the
historical behavior of consignees and constitute the terminal’s current
decision-making approach. Across evaluation periods, these heuristics exhibit
modest predictive performance: for $k = 300$, both achieve an average precision
of approximately 25\% and a recall close to 30\%, the latter being comparable to
the baseline rate. While these rules provide a useful starting point, the
results indicate clear limitations in their ability to prioritize containers
effectively.

Building on these baselines, the next stage complements the operational rules
with simple machine learning models, including decision trees and logistic
regression, denoted as \texttt{Baseline + ML1} and \texttt{Baseline + ML2}. These
hybrid approaches consistently improve performance, yielding gains of
approximately 10–15 percentage points in both precision and recall relative to
the rule-based baselines.

Finally, more sophisticated algorithms combined with specialized feature
engineering and a comprehensive use of available data result in substantial
performance improvements. In certain validation periods, the best-performing
model achieves precision levels of up to 75\% and recall values reaching 100\%,
indicating that all containers requiring service are correctly identified
within the prioritized set. This model, depicted by the purple line in
Figure~\ref{fig:serv}, corresponds to a \emph{Random Forest} classifier that
integrates multiple dimensions of information, including historical service
patterns, consignee characteristics, and recent behavioral indicators.

\subsubsection{Effect of Available Space in the Service Area}

\begin{figure}
    \centering
    \includegraphics[width=.9\textwidth]{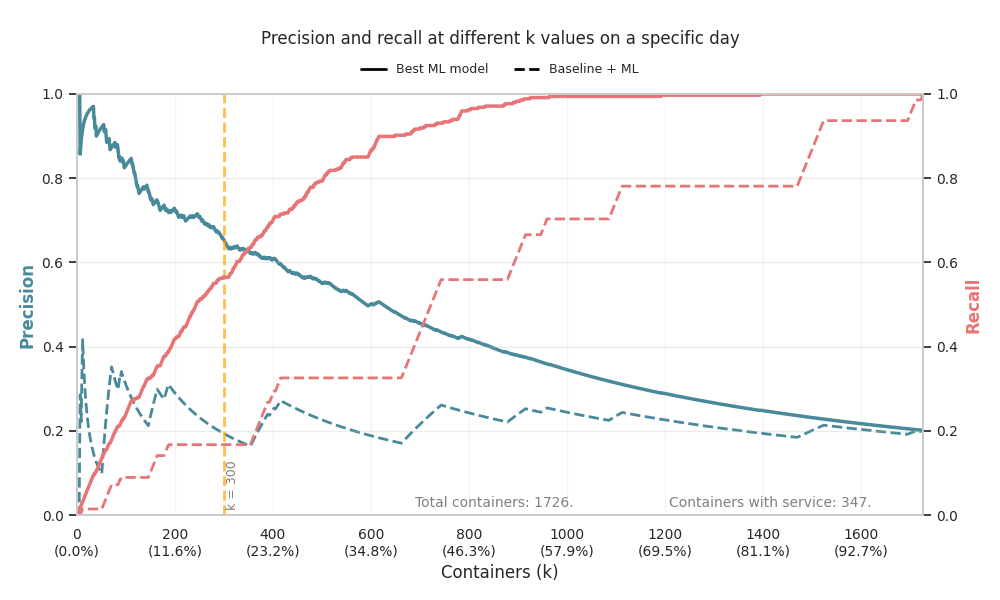}
    \caption{Precision and recall for different $k$ values under varying service area capacities.}
\label{fig:servrayid}
\end{figure}

Figure~\ref{fig:servrayid} illustrates how the best model would perform under a realistic operational scenario. On a typical day within the evaluation period, 1,726 containers arrived at the terminal, of which 347 required service. With $k = 300$, the top-performing model correctly identified 229 containers (66\% precision) and captured 201 of the 347 that actually required service (58\% recall).

This represents a substantial improvement—nearly 45 percentage points—over the
best-performing combination of baseline and simple ML models, which achieved
only 20\% precision and 15\% recall. A consistent pattern was observed across
experiments: as $k$ increases (i.e., more containers are prioritized), recall
improves—since more true positives are captured—but precision decreases due to a
higher number of false positives. This trade-off between precision and recall is
crucial for defining operational strategies according to space availability in
the service area.

\begin{figure}
    \centering
    \includegraphics[width=.9\textwidth]{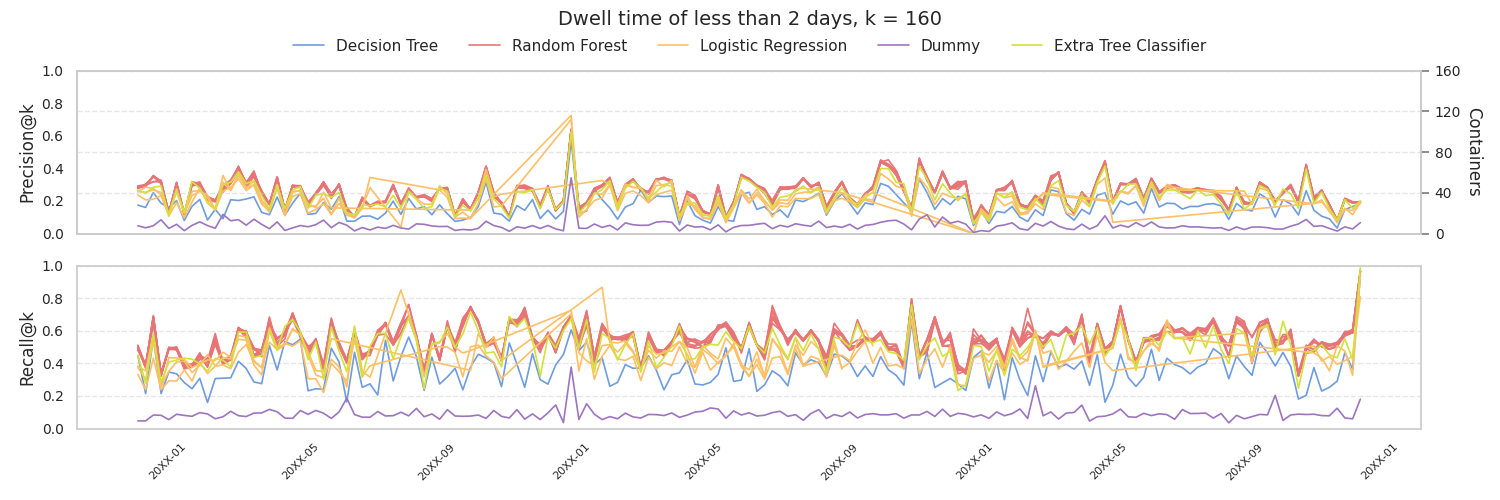}
    \caption{Weekly average precision and recall for dwell times of less than two days.}
    \label{fig:menos2}
\end{figure}
\begin{figure}
    \centering
    \includegraphics[width=.9\textwidth]{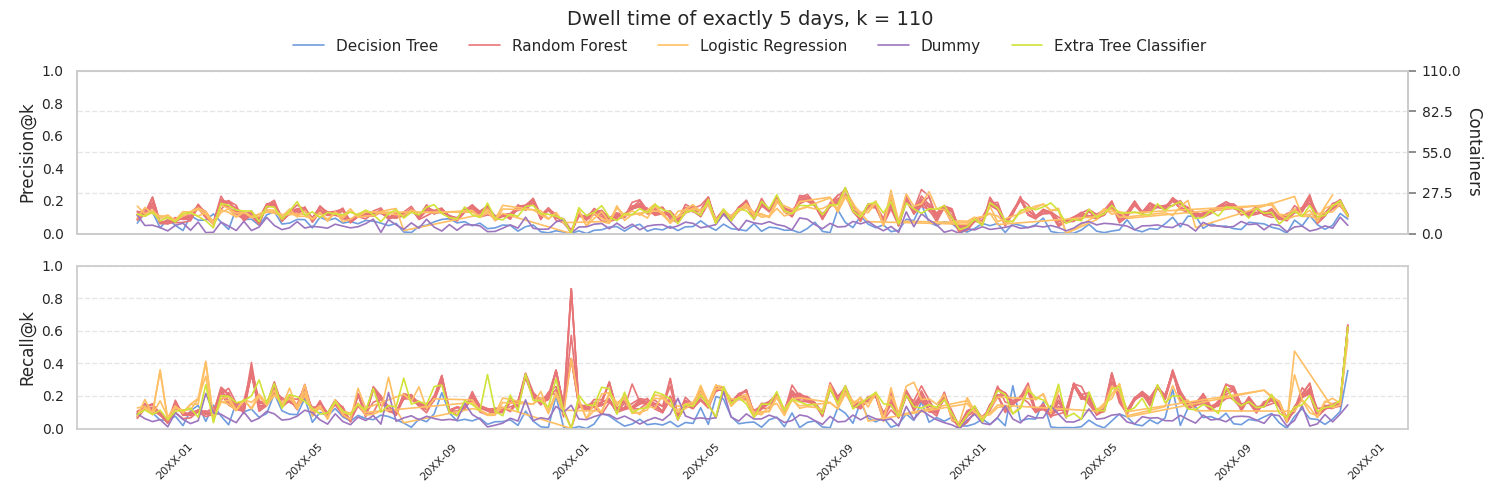}
   \caption{Weekly average precision and recall for containers with a dwell time of exactly five days. This label is used as a representative case for intermediate dwell-time categories (three, four, six, and seven days).}
    \label{fig:te5}
\end{figure}
\begin{figure}
    \centering
    \includegraphics[width=.9\textwidth]{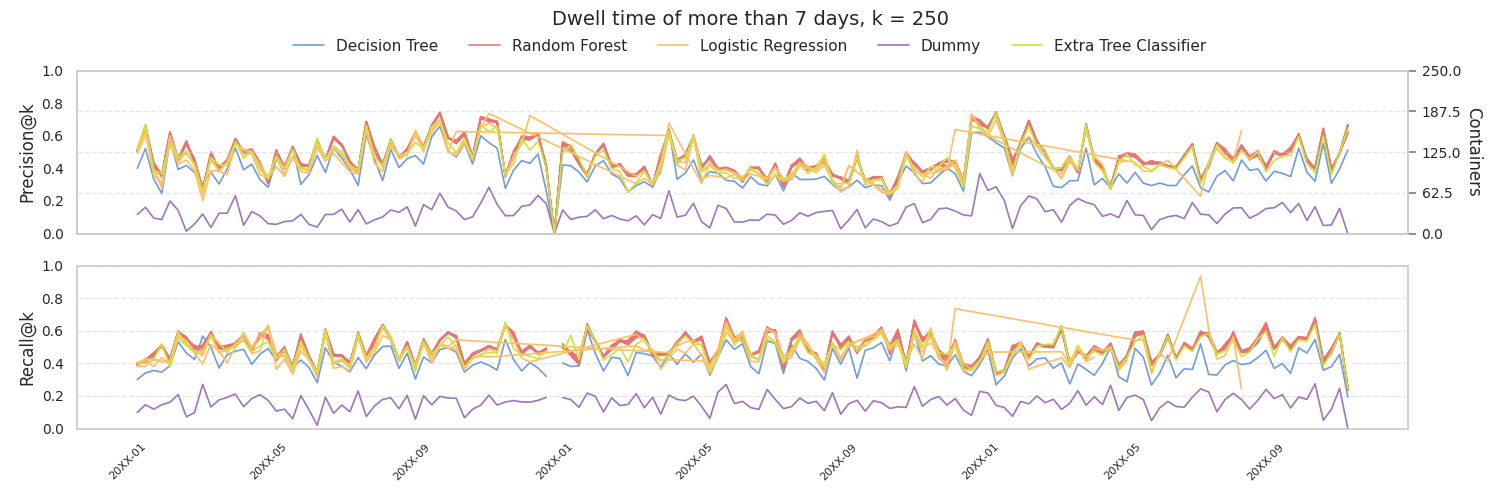}
    \caption{Weekly average precision and recall for dwell times longer than seven days.}
    \label{fig:mas7}
\end{figure}

\subsection{Dwell Time}\label{sec:dwelltime}

For dwell-time analysis, predictive models were trained for eight distinct
labels, each corresponding to a specific duration that a container may remain in
the terminal. This section reports results for three representative labels:
containers with dwell times of less than two days, exactly five days, and more
than seven days. The five-day label is selected as representative of the
intermediate dwell-time categories, which include containers staying exactly
three, four, six, and seven days.

Temporal evaluations for each label were conducted using an absolute $k$ value
corresponding to the historical number of containers that exhibited that
specific dwell time. This value was computed by weighting the baseline rate of
each label by the average number of containers arriving daily at the terminal (see Table~\ref{tab:baserate}).
Although the results shown in the figures reflect this label-specific approach,
it is important to note that the final dwell-time assignment for each container
is not determined independently. As detailed in Subsection~\ref{sec:decision},
this assignment is made through a decision rule that leverages optimal points
derived from ROC curves to select the most appropriate label for each case.

\subsubsection{Predictive Performance Across Dwell Time Labels}

Figures~\ref{fig:menos2}, \ref{fig:te5}, and \ref{fig:mas7} illustrate predictive
performance across selected dwell-time labels, revealing a clear pattern: model
performance is strongest at the temporal extremes. Labels corresponding to
containers that remain in the terminal for less than two days or more than seven
days exhibit substantially higher precision and recall than intermediate dwell
time categories. This suggests that operational behaviors associated with very
short or very long stays are more regular and therefore easier for predictive
models to capture.

The highest performance is observed for the label representing dwell times
longer than seven days (Figure~\ref{fig:mas7}). For this category, several models
achieve precision levels approaching 80\% and recall values exceeding 90\%,
outperforming the random baseline by more than 40 percentage points. These
results indicate that containers associated with prolonged stays follow more
consistent operational patterns, making them particularly amenable to
prediction.

For containers with dwell times of less than two days (Figure~\ref{fig:menos2}),
the best-performing models achieve precision levels between 30\% and 40\% and
recall values of up to 80\%, outperforming the random baseline by approximately
50 percentage points. This result highlights the usefulness of predictive
modeling for anticipating rapid departures and supporting the allocation of
easily accessible yard positions.

In contrast, labels representing intermediate dwell durations—ranging from two
to six days—exhibit more modest performance. For the representative intermediate
case corresponding to a dwell time of five days (Figure~\ref{fig:te5}),
predictive performance remains moderate. Most models, including
\textit{ExtraTreesClassifier}, \textit{Random Forest}, and
\textit{Logistic Regression}, achieve precision below 25\% and recall
rarely exceeding 40\%. Although all models consistently outperform the
random (\emph{dummy}) baseline, the limited gains suggest higher operational
variability within these temporal ranges, where multiple factors jointly
influence mid-range dwell times and hinder consistent pattern identification.

Overall, the results demonstrate that dwell-time prediction is most effective at
the extremes of the temporal spectrum, while intermediate durations remain more
challenging. From an operational perspective, this finding presents a clear
opportunity: containers predicted to remain in the terminal for extended periods
can be proactively assigned to lower-priority yard locations where early
retrieval is unlikely, thereby improving spatial efficiency and reducing
unnecessary container movements.

\subsubsection{Average Predictive Performance Over Time}

\begin{figure}
\centering
\includegraphics[width=1\textwidth]{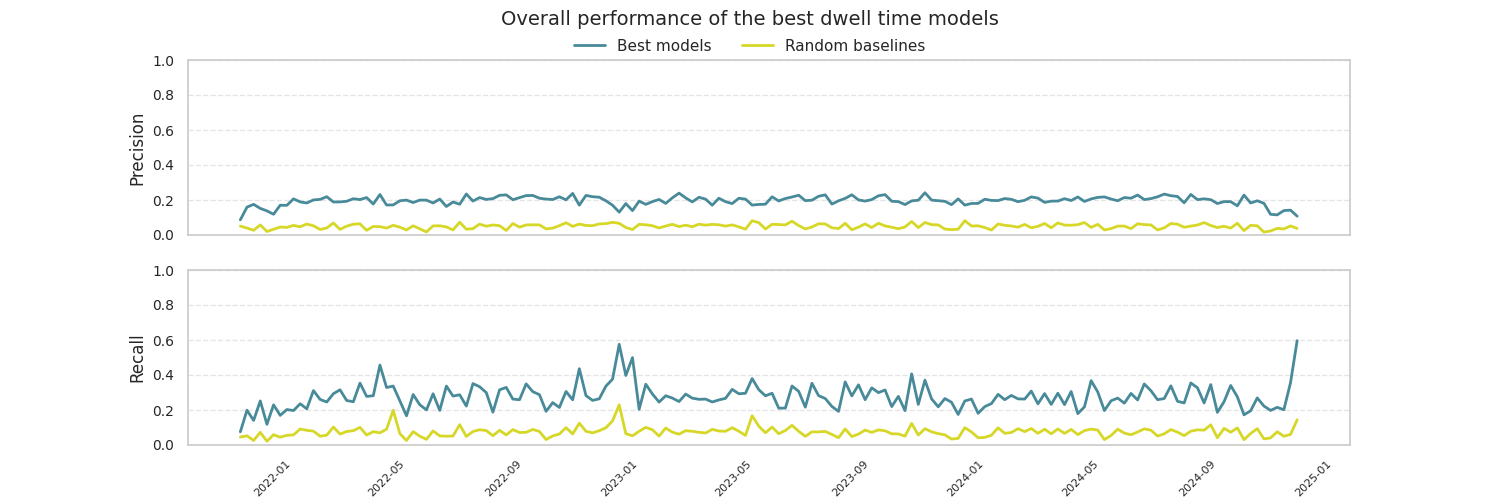}
\caption{Average precision and recall of the best-performing models across dwell-time labels.}
\label{fig:dia_te}
\end{figure}

Figure~\ref{fig:dia_te} shows the average precision and recall across all dwell
time labels. This analysis was performed by selecting the best-performing model
for each label, calculating its daily precision and recall, and then averaging
these values at a weekly level.

Results were compared against a non-predictive benchmark, representing a purely
random container assignment strategy. Across more than three years of validation
using real operational data, the developed models consistently outperformed the
random baseline in both precision and recall. Precision remained above 20\% in most
periods, while the random strategy barely reached 10\%. Recall doubled or even
tripled the levels achieved under random assignment.

These findings confirm that dwell-time prediction constitutes a practical
decision-support tool for terminal operations. Rather than reacting as
containers accumulate in the yard, operators can proactively plan and allocate
strategic positions from the outset, thereby reducing unnecessary movements and
optimizing spatial utilization.

\subsubsection{Operational Performance of Dwell Time Assignment}

\begin{figure}
\centering
\includegraphics[width=0.6\textwidth]{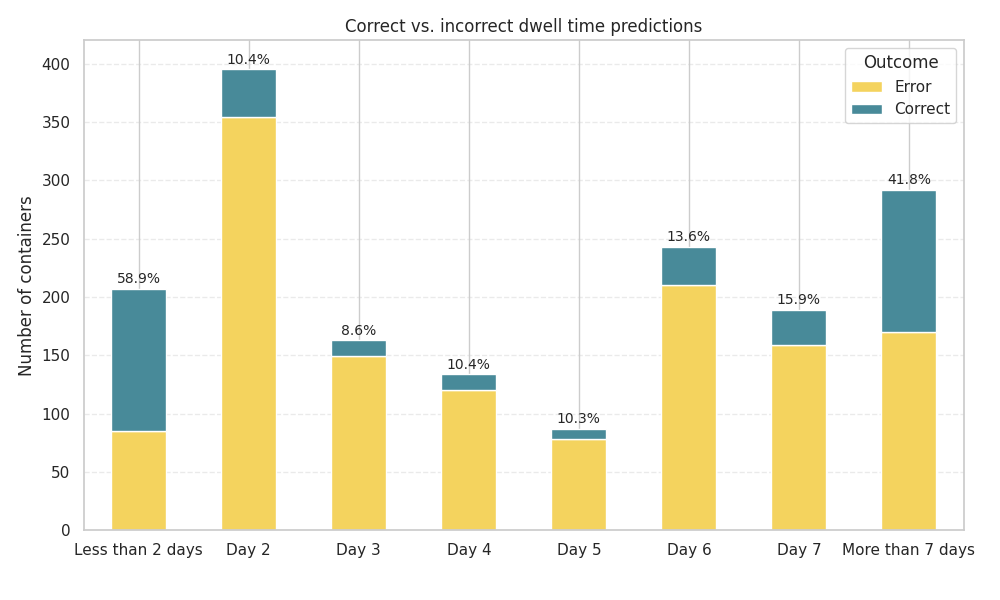}
\caption{Correct and incorrect classifications by dwell-time category on a typical operational day.}
\label{fig:globalte}
\end{figure}

Applying the decision rule described in Section~\ref{sec:decision}, each
container is assigned a unique dwell-time label. Figure~\ref{fig:globalte}
presents the distribution of correct and incorrect predictions relative to the
actual dwell times of containers on a typical operational day. This
visualization highlights the dwell-time ranges where models perform best and
those where misclassifications are concentrated.

Two strategic findings emerge from this analysis:

\begin{itemize}
    \item \textbf{Short-stay containers:} Models perform best at the extremes of operational behavior. For containers with dwell times shorter than two days, precision exceeds 58\%, representing a direct opportunity to assign these containers to easily accessible yard positions upon arrival.
    \item \textbf{Long-stay containers:} For containers staying more than seven days, the model correctly classifies approximately 42\% of cases, enabling early allocation to low-priority yard areas where prompt retrieval is not required.
\end{itemize}

These two groups—rapid-turnaround and long-stay containers—represent the greatest operational gains, as accurate identification significantly reduces unnecessary container movements.

On the other hand, \textbf{intermediate dwell times} (particularly between days 3 and 6) reveal areas for improvement. Precision declines and errors cluster in this range, likely due to greater operational variability and the heterogeneous factors influencing mid-range dwell durations. These results suggest that incorporating additional variables or modeling approaches may be necessary to better capture these intermediate behaviors.

\section{Discussion}

\subsection{Interpretation of Results in Context}

The predictive performance observed in this study aligns with and extends
findings from prior research on container dwell-time prediction.
\citet{kourouniotiDevelopmentModelsPredicting2016} reported that ANNs could
effectively classify containers by dwell time using port of origin, container
characteristics, and temporal features. Our results confirm these determinants
while demonstrating that ensemble methods—particularly Random Forest—achieve
superior performance when combined with consignee-level behavioral features and
rigorous temporal validation.

The finding that extreme dwell-time categories (less than two days and more than
seven days) are more predictable than intermediate durations reflects a pattern
consistent with operational reality. Short-stay containers typically correspond
to established logistics chains with reliable pickup schedules, while long-stay
containers often involve pre-clearance holds or storage arrangements that follow
identifiable patterns. Intermediate durations, by contrast, are subject to
greater variability from factors outside the terminal's visibility—such as
consignee decisions, documentation delays, or transportation availability.

\subsection{Comparison with existing approaches}

The comparison against operational baselines rather than random classifiers
distinguishes this study from much of the existing literature. While
\citet{yoonComparativeStudyMachine2023} and
\citet{gaeteg.DwellTimebasedContainer2017} demonstrated improvements over dummy
classifiers, our evaluation against the terminal's actual decision heuristics
provides a more operationally meaningful benchmark. The 45 percentage point
improvement in service prediction over combined baseline-plus-simple-ML
approaches represents a substantive gain with direct implications for daily
operations.

While container dwell time, yard congestion, and even customs-related
predictions have received considerable attention in the literature, the specific
problem of predicting which containers will require pre-clearance service from
the terminal's operational perspective has received limited attention. This gap
is consequential: given that 51\% of unproductive moves in this terminal are
associated with service-requiring containers, accurate prediction of this label
represents a direct lever for reducing operational inefficiency. Our model
addresses this gap by learning from the terminal's historical operational data
to anticipate service requirements in advance, enabling yard planners to
incorporate these predictions into their decisions before containers are
positioned in the yard.

Our treatment of cargo descriptions also relates to recent work by
\citet{xiePredictingOutterminalsImported2025}, who incorporated unstructured
cargo text as features for predicting container exit terminals. While their
approach used raw text processing techniques, we adopted a structured
alternative: mapping free-text cargo descriptions to standardized HS
(Harmonized System) chapters and sections through classification and record
linkage. This structured representation reduces dimensionality while preserving
semantically meaningful cargo categories, and enables the construction of
interpretable behavioral features (e.g., historical service rates by cargo
chapter) that would be difficult to derive from raw text embeddings.

\subsection{Methodological Considerations}

The use of temporal cross-validation
\cite[see][]{robertsCrossvalidationStrategiesData2017} addresses a common
methodological pitfall in port logistics studies. Many predictive studies employ
conventional cross-validation, which can lead to data leakage when temporal
dependencies exist in the data. By training exclusively on past data and
evaluating on future periods, our approach ensures that reported performance
metrics reflect realistic deployment conditions.

The decision rule combining eight binary classifiers through ROC-derived
thresholds and score ranking offers a practical solution to the multi-class
dwell-time prediction problem. While a single multi-class classifier might
appear more elegant, the binary decomposition allows for label-specific
optimization and provides interpretable confidence measures for each prediction.
The Jaccard similarity analysis (see Appendix~\ref{app:jaccard}) confirms that different labels identify largely
distinct container subsets, validating this approach.

\subsection{Limitations}

Several limitations should be acknowledged. First, the models rely on
information available at the time of vessel arrival, excluding factors that may
emerge during the container's stay—such as documentation issues or consignee
requests for extended storage.

Second, this study was conducted at a single terminal in Mexico. While the
methodology is transferable, the specific feature importance rankings and
optimal hyperparameters may vary across terminals with different operational
characteristics, cargo mixes, or regulatory contexts. Notably, the two
prediction tasks may exhibit different generalization patterns. Service
prediction depends on pre-clearance processes and operational practices that
vary across contexts, suggesting that models trained in one setting may require
significant recalibration elsewhere. Dwell-time prediction, by
contrast, captures logistics chain dynamics—short-stay containers reflecting
efficient pickup operations and long-stay containers reflecting storage
arrangements or holds—that likely share structural similarities across
terminals, potentially enabling better cross-terminal transfer of learned
patterns for extreme dwell-time categories.

Finally, this study validates predictive performance rather than directly
measuring operational outcomes such as reshuffle reduction. Translating
predictive accuracy into yard efficiency gains requires operational deployment
or simulation studies with stacking position and retrieval sequence data not
available in the current study.

\subsection{Indicative Estimates of Potential Reshuffle Reduction}

The potential operational impact of the predictive models can be estimated
through a framework that connects prediction quality to reshuffle reduction.

\subsubsection{Aggregate Impact Estimate}

Terminal statistics indicate that 75\% of container moves are classified as
unproductive, and 51\% of these involve containers requiring service.
Let $\alpha = 0.75$ denote the unproductive move ratio and $\beta = 0.51$ the
fraction attributable to service-requiring containers. Let $\delta_s$ represent the
fraction of service-related reshuffles that could be avoided through
prediction-informed placement, and $\delta_d$ the corresponding fraction for
dwell-time-informed stacking of non-service containers.

The reduction in total handling operations is bounded by:
\begin{equation}
\Delta_{\text{total}} = \alpha \cdot \beta \cdot \delta_s + \alpha \cdot (1 - \beta) \cdot \delta_d
\end{equation}

For service prediction, observed gains of approximately 50 percentage points in
recall over the operational baseline suggest $\delta_s \in [0.25, 0.40]$ under
reasonable stacking policies. For dwell-time prediction, the strong performance
at temporal extremes---precision exceeding 58\% for short-stay and 42\% for
long-stay containers, which together represent approximately 36\% of all
containers---suggests $\delta_d \in [0.10, 0.20]$. Under these assumptions, the
estimated reduction in total handling operations lies between \textbf{13\% and
23\%}, depending on stacking policy sophistication. 

\subsubsection{Disaggregated Analysis by Container Category}

The impact is not uniformly distributed across container categories:

\paragraph{Service containers} (33\% of arrivals, 51\% of unproductive moves)
have outsized operational impact because service events create
\emph{discontinuities} in the container trajectory---the container must be
extracted and moved to the service area regardless of its yard position. The
model's recall of up to 100\% at $k=300$ suggests that under sufficient service
area capacity, nearly all service-related reshuffles could be addressed through
prediction-informed placement.

\paragraph{Short-stay containers} ($<$2 days, $\sim$8\% of arrivals) have the
highest per-unit reshuffle potential because they depart before most containers
stacked around them. With model precision exceeding 58\%, prediction-informed
placement in accessible positions directly eliminates these high-cost
reshuffles.

\paragraph{Long-stay containers} ($>$7 days, $\sim$28\% of arrivals) present the
inverse opportunity: correctly identified long-stay containers can be placed in
deep storage positions where they will not obstruct earlier departures.

\paragraph{Intermediate-stay containers} (2--7 days, $\sim$64\% of arrivals)
contribute less marginal value per prediction due to lower prediction quality
and smaller operational cost of adjacent-day misclassification. Nevertheless,
even coarse classification provides stacking information superior to random
assignment.

\subsubsection{Stacking Policy Sensitivity and Asymmetric Costs}

The magnitude of reshuffle reduction depends on the stacking policy that
consumes the predictions. Simple zone segregation---routing containers to
different yard areas by predicted category---requires minimal system integration
and captures 10--15\% reduction in total moves. More sophisticated within-block
ordering, placing predicted earlier-departure containers atop later-departure
containers, yields 15--25\% reduction but requires integration with yard
management systems.

Prediction errors carry asymmetric operational costs. Misclassifying a service
container as non-service (false negative) incurs high cost: the container is
buried in a standard block and must be excavated for service. The reverse
error merely wastes premium service-area space. Similarly, misclassifying a
short-stay container as long-stay generates multiple reshuffles at early
departure, whereas the inverse error wastes accessible positions but causes no
cascade. This asymmetry suggests that for service prediction, higher recall
should be prioritized over higher precision, consistent with the operational
capacity constraint ($k = 300$) employed in this study.

\subsubsection{Validation Framework}

While the analytical estimates above provide indicative bounds, definitive
quantification requires simulation using historical operational data. The
movements dataset (13 million records) enables reconstruction of actual stacking
configurations and retrieval reshuffles. A counterfactual simulation comparing
historical placements against prediction-informed placements---under physical
constraints and using the model's actual predictions including errors---would
quantify achievable reshuffle reduction. Such simulation, together with
perfect-information upper bounds establishing theoretical maxima, is the subject
of ongoing work.

\subsection{Theoretical Contributions}

This study contributes to the growing literature on prediction-optimization
integration in port logistics \citep{jahangardLeveragingMachineLearning2025}. By
demonstrating that predictive outputs can meaningfully inform stacking
decisions, we provide empirical support for data-driven approaches to yard
management. The combination of service prediction and dwell-time estimation
within a unified framework offers a more comprehensive decision-support
capability than either prediction alone.

\section{Conclusion}

This study developed and validated a machine learning framework for predicting
container service requirements and dwell times at a Mexican port terminal. The
framework substantially outperforms current operational heuristics: service
prediction achieved double the precision and recall of baseline approaches,
while dwell-time prediction demonstrated strong accuracy for operationally
critical categories—short-stay containers requiring accessible placement and
long-stay containers suitable for storage areas.

The methodology employs rigorous temporal cross-validation, ensuring that
reported metrics reflect realistic deployment conditions. The combination of
service and dwell-time predictions
within a unified framework provides comprehensive decision support for yard
planners, enabling informed stacking decisions at the time of container arrival.

While predictive performance is validated, translating these gains into measured
reshuffle reductions requires operational deployment or simulation studies.
Several directions merit further investigation. First, operational validation
through simulation studies would strengthen the connection between predictive
accuracy and yard efficiency. Using historical retrieval sequences, a backtest
could compare reshuffle counts under current stacking policies versus
dwell-informed placement rules \citep[cf.][]{gaeteg.DwellTimebasedContainer2017}. Such analysis requires detailed stacking position and retrieval sequence data,
which cannot be fully reconstructed from the available movement records in the
current study, but would provide direct quantification of move reductions
achievable through prediction-informed placement.

Second, the model could be extended to incorporate real-time information
updates as containers progress through their dwell period, potentially
improving predictions for intermediate dwell-time categories. Third,
integration with vessel scheduling systems could enable proactive yard
reorganization before peak retrieval periods. Finally, extending the framework
to other terminal operations—such as equipment allocation and gate
scheduling—represents a natural progression toward comprehensive data-driven
terminal management.

\section*{Acknowledgments}

The authors thank Zaid Hernández Solano, Carlos Eduardo Olvera Azuara, and
Roberto Villarreal Ramírez for their contributions to data processing and analysis during some stages of this project.

\section*{Author Contributions}

\textbf{Elena Villalobos}: Conceptualization, Methodology, Software, Formal
analysis, Investigation, Data Curation, Writing -- Original Draft, Writing -- Review \& Editing, Visualization.
\textbf{Adolfo De Unánue}: Conceptualization, Methodology, Supervision,
Software, Formal analysis, Investigation, Writing -- Review \& Editing.
\textbf{Fernanda Sobrino}: Investigation, Writing -- Review \&
Editing. \textbf{David Aké}: Project administration. \textbf{Stephany Cisneros}:
Investigation. \textbf{Jorge Lecona}: Resources, Project administration, Writing
-- Review \& Editing. \textbf{Alejandra Matadamaz}: Domain expertise, Writing --
Review \& Editing.

\section*{Declaration of Competing Interests}

Jorge Lecona and Alejandra Matadamaz are employees of the container terminal where this study was conducted. The study design, analysis,
and interpretation were conducted independently by the academic authors.

\appendix
\section{Table of evaluated algorithms and hyperparameters. }

\begin{table}[htbp]
\centering
\caption{Summary of evaluated algorithms and hyperparameter configurations.}
\begin{tabular}{lll}
\toprule
\textbf{Algorithm} & \textbf{Hyperparameter} & \textbf{Values tested} \\
\midrule
DecisionTreeClassifier & criterion & \texttt{gini} \\
 & max\_depth & 5, 10, 50 \\
 & min\_samples\_split & 10, 50, 100 \\
\midrule
RandomForestClassifier & n\_estimators & 200, 300 \\
 & criterion & \texttt{gini} \\
 & max\_depth & 5, 10 \\
 & max\_features & \texttt{sqrt} \\
 & min\_samples\_split & 10, 50 \\
\midrule
ScaledLogisticRegression & penalty & \texttt{l1}, \texttt{l2} \\
 & C & 0.0001, 0.01, 0.1, 1.0 \\
\midrule
ExtraTreesClassifier & n\_estimators & 500 \\
 & criterion & \texttt{gini} \\
 & max\_depth & 5, 10 \\
 & max\_features & \texttt{sqrt} \\
 & min\_samples\_split & 50, 100 \\
\bottomrule
\end{tabular}
\label{app:dwell}
\end{table}

\section{Similarity Analysis Across Dwell Time Labels}\label{app:jaccard}

For each container, the model generates predictions across the eight possible
dwell-time labels. To determine whether models prioritize the same containers
across different labels or identify distinct subsets, we computed the pairwise
similarity between the sets of containers prioritized by each label using the
Jaccard similarity coefficient, which measures the proportion of shared elements
between two sets. A value close to 1 indicates high overlap (the same containers
appear in both labels), whereas a value close to 0 indicates distinct subsets.

\begin{figure}[htbp]
\centering
\includegraphics[width=0.6\textwidth]{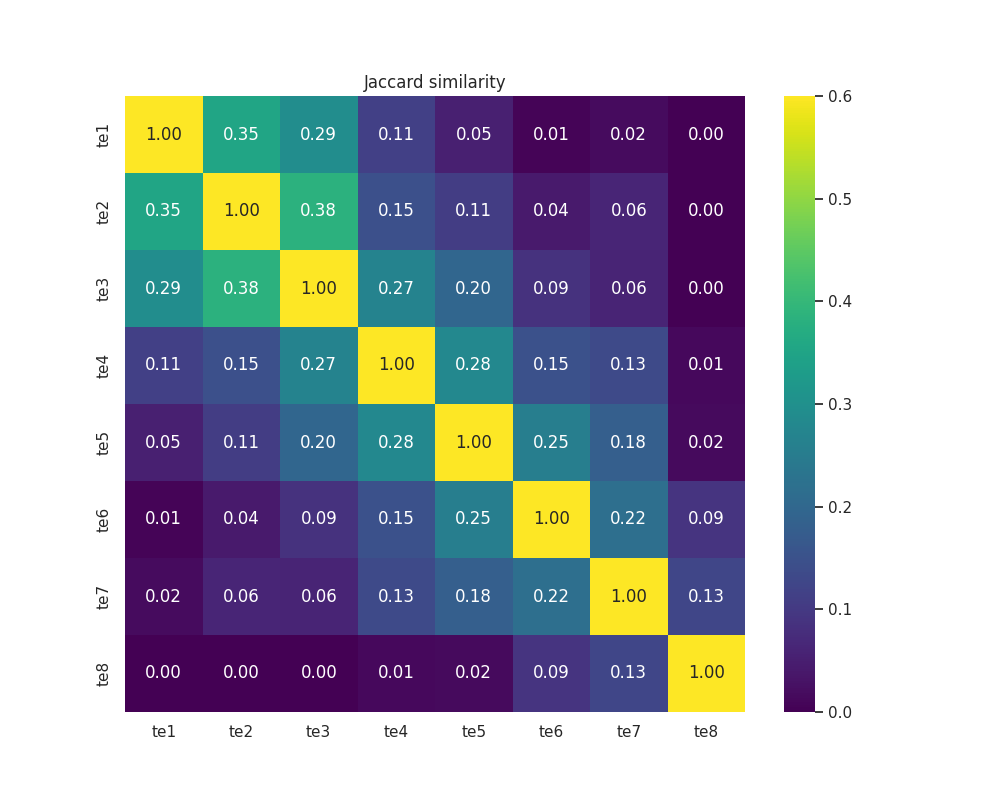}
\caption{Jaccard similarity matrix across dwell-time labels.}
\label{fig:jac_sim}
\end{figure}

Figure~\ref{fig:jac_sim} shows the Jaccard similarity matrix across all
combinations of dwell-time labels. Higher similarity values are concentrated
between adjacent dwell-time categories—e.g., between the two-day label
(\texttt{te2}) and the three-day label (\texttt{te3}) (0.38), or between the
less-than-two-days label (\texttt{te1}) and the two-day label (\texttt{te2})
(0.35)—indicating partial overlap in prioritized containers for these models.
However, for most other combinations, similarity is substantially lower,
suggesting that the models identify largely distinct container subsets. This
provides evidence that the models are learning differentiated and label-specific
patterns rather than replicating identical predictions across multiple labels.

\bibliographystyle{unsrtnat}
\bibliography{references}



\end{document}